%% file: root.tex
\documentclass[letterpaper, 10 pt, conference]{ieeeconf} 
\IEEEoverridecommandlockouts

\usepackage[T1]{fontenc}
\usepackage{amsmath,amsfonts}
\usepackage{cite}
\usepackage{stfloats}
\usepackage{graphicx}
\usepackage[caption=false,font=footnotesize]{subfig}
\usepackage{url}
\usepackage{hyperref}
\usepackage{tabularx}
\usepackage{siunitx}
\usepackage{booktabs}

\usepackage{tikz}
\usetikzlibrary{arrows.meta}
\usetikzlibrary{arrows}
\usepackage{pgfplots}
\usepgfplotslibrary{statistics}
\pgfplotsset{compat=1.18}

\usepackage{xcolor}
\definecolor{Black}{HTML}{000000}
\definecolor{Blue}{HTML}{0065bd}
\definecolor{Bluelight}{HTML}{D6E8F7}
\definecolor{Red}{HTML}{8C000F}
\definecolor{Orange}{HTML}{E97132}
\definecolor{Green}{HTML}{A2AD00}
\definecolor{GreenCR}{HTML}{008000}
\definecolor{LightGray}{HTML}{e7e7e7}

\newcommand{\refpath}{
  \tikz{
    \draw[line width = 1.75pt, Black] (0,0) -- (0.4,0);
    \node[inner sep=0pt] at (0,-0.05) {};
  }
}

\title{\LARGE \bf
Towards Safe Autonomous Driving: A Real-Time Safeguarding Concept for Motion Planning Algorithms
}

\author{Korbinian Moller, Rafael Neher, Marvin Seegert, Johannes Betz%
\thanks{K. Moller, R. Neher, M. Seegert, and J. Betz are with the Professorship of Autonomous Vehicle Systems, TUM School of Engineering and Design, Technical University of Munich, 85748 Garching, Germany; Munich Institute of Robotics and Machine Intelligence (MIRMI).}%
}

\begin{document}

\maketitle

%%%%%%%%%%%%%%%%%%%%%%%%%%%%%%%%%%%%%%%%%%%%%%%%%%%%%%%%%
%%% Abstract
%%%%%%%%%%%%%%%%%%%%%%%%%%%%%%%%%%%%%%%%%%%%%%%%%%%%%%%%%

\begin{abstract}

Ensuring the functional safety of motion planning modules in autonomous vehicles remains a critical challenge, especially when dealing with complex or learning-based software. Online verification has emerged as a promising approach to monitor such systems at runtime, yet its integration into embedded real-time environments remains limited. This work presents a safeguarding concept for motion planning that extends prior approaches by introducing a time safeguard. While existing methods focus on geometric and dynamic feasibility, our approach additionally monitors the temporal consistency of planning outputs to ensure timely system response. A prototypical implementation on a real-time operating system evaluates trajectory candidates using constraint-based feasibility checks and cost-based plausibility metrics. Preliminary results show that the safeguarding module operates within real-time bounds and effectively detects unsafe trajectories. However, the full integration of the time safeguard logic and fallback strategies is ongoing. This study contributes a modular and extensible framework for runtime trajectory verification and highlights key aspects for deployment on automotive-grade hardware. Future work includes completing the safeguarding logic and validating its effectiveness through hardware-in-the-loop simulations and vehicle-based testing. The code is available at: \url{https://github.com/TUM-AVS/motion-planning-supervisor}.

\end{abstract}

%%%%%%%%%%%%%%%%%%%%%%%%%%%%%%%%%%%%%%%%%%%%%%%%%%%%%%%%%
%%% Keywords
%%%%%%%%%%%%%%%%%%%%%%%%%%%%%%%%%%%%%%%%%%%%%%%%%%%%%%%%%
\vspace{0.1cm}

%%%%%%%%%%%%%%%%%%%%%%%%%%%%%%%%%%%%%%%%%%%%%%%%%%%%%%%%%
%%% Introduction
%%%%%%%%%%%%%%%%%%%%%%%%%%%%%%%%%%%%%%%%%%%%%%%%%%%%%%%%%

\section{Introduction}
\label{sec:introduction}

In a world shaped by technology, autonomous vehicles (AVs) are expected to revolutionize mobility by enhancing traffic efficiency and reducing accidents~\cite{Bobisse2019}. AV systems rely on complex algorithms~\cite{Pendleton2017} that must continuously interpret their environment and make real-time decisions to ensure both safety and efficiency. However, despite significant advancements, practical experience and collision reports have highlighted persistent challenges that must be addressed before AVs can be safely deployed at scale~\cite{Pokorny2022}.

As AVs expand their operational design domain (ODD) from controlled environments such as highways to complex urban areas, ensuring safe and reliable operation becomes increasingly difficult~\cite{Taxonomy2024}. Modern motion planners often use heuristic or learning-based algorithms~\cite{Paden2016, Katrakazas2015}, which are powerful, yet non-transparent and difficult to verify~\cite{Koopman2016}. Classical validation methods fall short in these cases, particularly regarding the functional safety requirements of ISO~26262~\cite{ISO26262}.

\begin{figure}[!t]
    \centering
    \includegraphics[width=\linewidth]{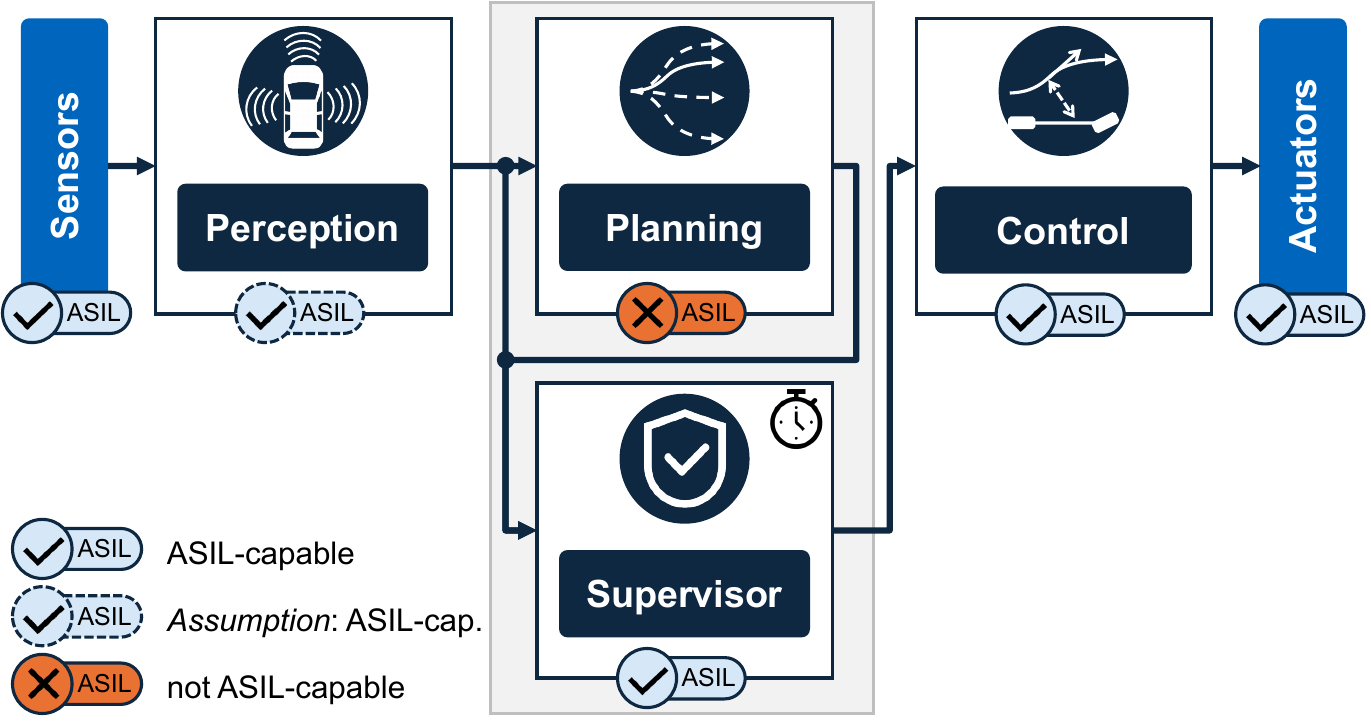}
    \caption{Conceptual ASIL decomposition for motion planning. An independent ASIL-rated Supervisor monitors a non-ASIL-capable planner online to enable safe system-level approval~\cite{Stahl2020}.}
    \label{fig:introduction}%
\end{figure}

To address this, online verification (OV) introduces an ASIL-rated Supervisor that monitors outputs from non-certified planning modules during runtime~\cite{Stahl2020, Stahl2021}. Such safeguarding enables modular system approval via ASIL decomposition (\autoref{fig:introduction}). Most approaches, however, focus on constraint violations (e.g., collisions) and neglect temporal gaps between valid outputs~\cite{Stahl2020, Stahl2021}.

To close this gap, we introduce the concept of a time safeguard, which tracks the temporal continuity of safe planning outputs. The approach is based on the observation that large intervals between successive valid trajectories pose a safety risk, as the system may lose responsiveness to rapid changes in the driving environment. While human perception-reaction times range from \SIrange{700}{1500}{\milli\second}~\cite{Green2000}, safety benchmarks for AVs define a maximum response time of \SI{100}{\milli\second}~\cite{Becker2020}. Ensuring this response capability requires tight integration of runtime validation and timing supervision within the planning-execution loop.

We present a modular safeguarding framework implemented on embedded real-time hardware, where trajectory candidates are verified using feasibility checks and cost-based plausibility metrics. While full-time safeguard logic and fallback mechanisms are still under development, the architecture is designed to support these features. The main contributions of this paper are:
\begin{itemize}
  \item An extended safeguarding concept for AV motion planning that incorporates temporal consistency checks.
  \item A prototypical real-time implementation of the safeguard on embedded real-time hardware.
  \item An open-source codebase to support reproducibility and further development in the research community.
\end{itemize}

%%%%%%%%%%%%%%%%%%%%%%%%%%%%%%%%%%%%%%%%%%%%%%%%%%%%%%%%%
%%% Related Work
%%%%%%%%%%%%%%%%%%%%%%%%%%%%%%%%%%%%%%%%%%%%%%%%%%%%%%%%%
\section{Related Work}
\label{sec:relatedwork}

While safety in automotive systems is formally addressed by ISO~26262~\cite{ISO26262} and ISO~21448~\cite{ISO21448}, their applicability is limited in the context of modern AV software~\cite{Monkhouse2017, Wu2020}. Both standards presuppose traceable architectures and deterministic behavior, which are often violated by heuristic or learning-based planning components~\cite{Kuznietsov2024}. For example,~\cite{Mandrioli2022} highlights that adaptive systems may change their behavior during operation, rendering static verification and certification approaches ineffective.

A range of verification approaches has been explored to complement traditional safety standards. Simulation-based testing has become a widely adopted strategy, with traffic-level validation~\cite{Kitajima2019} and scenario-based generation frameworks~\cite{Chang2024} enabling scalable evaluation of critical situations. Nonetheless, high-fidelity modeling remains challenging, particularly regarding sensor simulation and interactive agent behavior~\cite{Jang2024, Kanwischer2022}. Additional approaches include shadow mode testing, where an AV system passively observes real-world scenarios while a human driver remains in control~\cite{Wang2019-validation}. While these methods provide incremental improvements, they remain inherently limited in scalability and fail to comprehensively guarantee safety for highly complex, adaptive systems. Analytical techniques such as barrier certificates, temporal logic specifications, and safety-distance frameworks like RSS~\cite{Tuncali2018, Maierhofer2020, ShalevShwartz2017, Gao2023} offer formal safety guarantees under idealized assumptions. However, they often do not scale to complex, high-dimensional planning systems and may induce overly conservative behavior when applied to real-world AV systems.

In response to these limitations, online risk assessment methods have gained traction. They estimate probabilistic metrics such as collision risk or harm potential based on predicted agent behaviors and environmental uncertainty~\cite{Geisslinger2021, Zhang2020}. While useful for in-situ awareness and planning tradeoffs, these approaches do not enforce guarantees and are not externally certifiable.

A structurally different paradigm is online verification (OV)~\cite{Pek2020, Pek2021}, which introduces a runtime safety layer capable of evaluating motion plans before execution. The approach was conceptualized for autonomous racing by Stahl et al.~\cite{Stahl2020, Stahl2021}, who proposed a high-assurance Supervisor that monitors safety-relevant trajectory properties~\cite{Censi2019} to intercept unsafe outputs before execution. This architecture enables the use of uncertified planning components under the supervision of a formally verified ASIL-D monitor, allowing for system-level approval through ASIL decomposition~\cite{ISO26262}. Inspired by these academic foundations and reinforced by practical implementations~\cite{ARM2024}, OV is increasingly seen as a cost-effective alternative to full software redundancy.

However, a critical limitation remains: most OV implementations focus solely on functional correctness at runtime, without addressing the timing of verification itself. In safety-critical systems, not only must each planning output be valid, but it must also arrive within bounded intervals to ensure continuous responsiveness. If the delay between valid plans exceeds a critical threshold, the system may fail to react in time to environmental changes. Real-time systems research emphasizes that functional and timing correctness are equally essential for safety~\cite{Buttazzo2024}. Hard real-time systems are expected to meet strict deadlines without exception, while soft real-time systems may tolerate occasional violations. Addressing this shortcoming requires safeguarding architectures that are both formally defined and independently executed with real-time guarantees.

%%%%%%%%%%%%%%%%%%%%%%%%%%%%%%%%%%%%%%%%%%%%%%%%%%%%%%%%%
%%% Methodology
%%%%%%%%%%%%%%%%%%%%%%%%%%%%%%%%%%%%%%%%%%%%%%%%%%%%%%%%%
\section{Methodology}
\label{sec:method}

As shown in the previous section, most existing safeguarding approaches focus on functional correctness but do not address the timing of verification. In particular, they often lack dedicated execution environments and run as part of the main software stack. This is critical in safety-relevant applications, where missed deadlines can undermine monitoring effectiveness. To overcome this limitation, we propose a real-time safeguarding framework that evaluates motion planning outputs independently of the primary system. Our design explicitly considers execution timing and platform isolation to ensure reliable and certifiable behavior. 

\begin{figure*}[!t]
    \centering
    \includegraphics[width=0.85\linewidth]{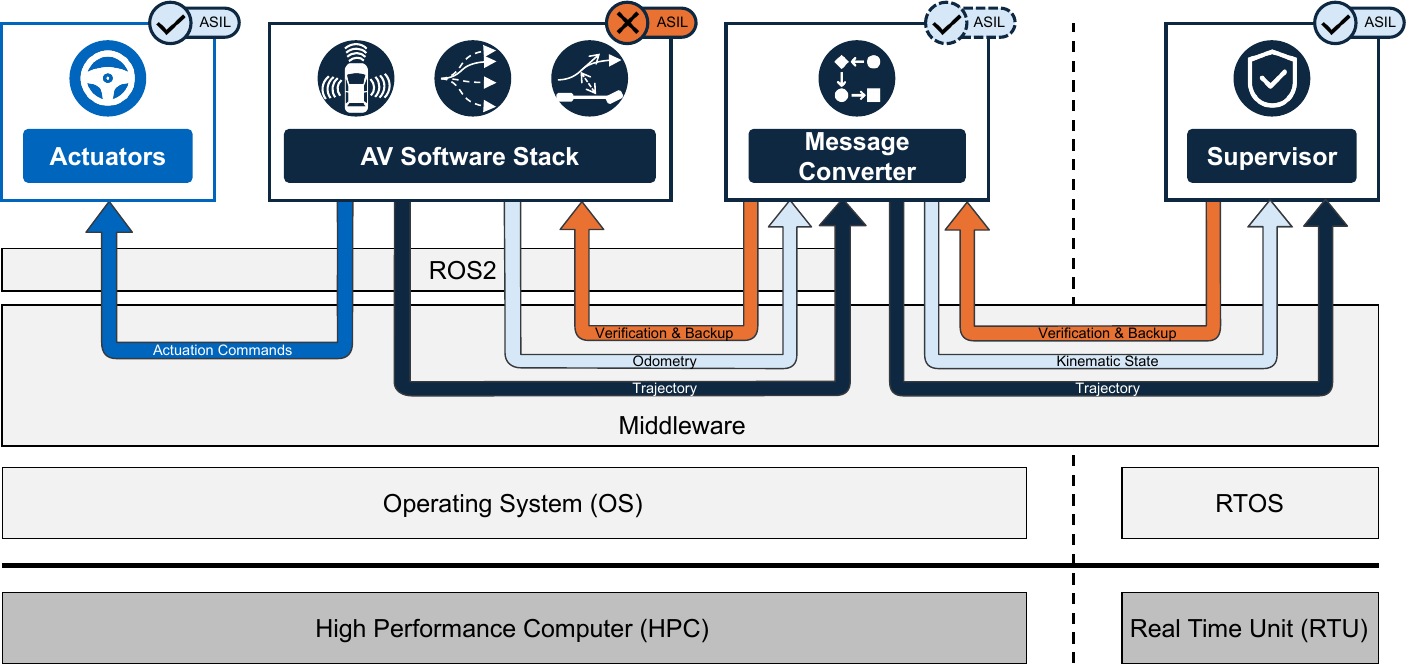}
    \caption{Overview of the proposed safeguarding architecture. The AV stack runs on a high-performance computer with a general-purpose operating system, while the Supervisor executes independently on a real-time unit (RTU) with RTOS support. Middleware communication is realized via ROS 2 and DDS.}
    \label{fig:methodology}%
\end{figure*}

Figure~\ref{fig:methodology} illustrates the overall architecture of the proposed safeguarding framework. We design the AV software stack to run on a high-performance computer (HPC) using a standard general-purpose operating system. This reflects common deployment patterns in prototypical autonomous vehicle platforms, where complex planning and perception pipelines require significant computational resources. The HPC generates motion plans and publishes them via middleware, such as ROS 2 and DDS, for further processing. In parallel, we deploy the Supervisor on a dedicated real-time unit (RTU), which runs a Real-time operating system (RTOS) to guarantee deterministic execution. By executing the Supervisor outside the AV stack and on separate hardware, we ensure architectural independence and minimize cross-component interference. The Supervisor receives each planned trajectory and evaluates it with respect to feasibility, timing, and plausibility. If the verification fails, the Supervisor can reject or override the trajectory before actuation.

\subsection{System Inputs and Assumptions}
\label{sec:inputs}

The Supervisor assumes no access to the internal state of the AV software stack and treats it as a black box. Instead, it evaluates observable outputs against a predefined set of safety criteria.

We denote the planned trajectory at time step \( t \) as
\begin{equation}
    \xi_t = \left\{ \mathbf{x}_0, \mathbf{x}_1, \dots, \mathbf{x}_N \right\},
\end{equation}
where each trajectory point \( \mathbf{x}_i \in \mathbb{R}^n \) represents the planned ego vehicle (EV) state at a future time step \( t+i \). In typical applications, this state vector contains position, velocity, acceleration, and orientation
\begin{equation}
\mathbf{x}_i = \left(x_i, y_i, v_i, a_i, \psi_i\right)^\top.
\label{eq:x_i}
\end{equation}
The trajectory is defined over a discrete time horizon \( [0, N] \), with a uniform time increment \( \Delta t \). The initial state \( \mathbf{x}_0 \) corresponds to the current state of the EV. The current pose estimate of the EV is required to be continuously available. In addition, the Supervisor requires access to the environmental context, e.g., surrounding dynamic objects. At each time step, we expect a set of detected and predicted agents,
\begin{equation}
\mathcal{O}_t = \left\{ o_t^1, o_t^2, \dots, o_t^M \right\},
\end{equation}
where each object \( o_t^j \) contains current and predicted future states of other traffic participants. We assume that object detection and prediction are performed upstream in the AV stack and that the resulting data are transmitted to the Supervisor via middleware.

\subsection{Communication Design}
\label{sec:communication}

Modern AV software stacks such as Autoware\footnote{https://autoware.org/} are commonly built on ROS~2 due to its modular structure, broad ecosystem support, and native integration with robotic frameworks. ROS~2 enables flexible development and message-passing via Data Distribution Service (DDS) middleware backends. However, it is not designed for safety-critical applications, as it lacks timing guarantees and exhibits significant overhead on constrained embedded platforms. RTOS such as Zephyr~\cite{Zephyr2025} typically do not support full ROS~2 stacks. Instead, they provide lightweight DDS clients or partial micro-ROS compatibility. In our concept, we adopt Cyclone DDS as middleware on the RTU, which offers deterministic performance and is compatible with Zephyr's real-time execution model.

To ensure compatibility with ROS~2-based planning stacks, we use a message conversion interface that translates selected topics into Cyclone DDS-compliant formats. This interface allows the Supervisor to receive data without executing any ROS~2-specific components. The converter subscribes to ROS~2 topics on the HPC and republishes them in a lightweight DDS message~\cite{ARM2024safetyisland}.

\subsection{Supervisor Execution Pipeline}
\label{sec:execution}

The Supervisor operates in a cyclic execution loop with deterministic timing. After initialization, it waits for a new trajectory message \( \xi_t \) within a defined maximum reaction time \( T_{\text{max}} \). Let \( t_{\text{recv}} \) denote the time of the most recent valid trajectory received. At each cycle time \( t \), the system checks whether
\begin{equation}
t - t_{\text{recv}} \leq T_{\text{max}}.
\end{equation}
If this condition is violated, i.e., no new trajectory is received within the specified window, the Supervisor concludes that a fault has occurred in the upstream system. Possible causes include invalid trajectory generation, planning timeouts, software faults, or hardware failure on the HPC.

If a valid trajectory \( \xi_t \) is received in time, the Supervisor evaluates it against a defined set of safety criteria, as detailed in Subsection~\ref{sec:evaluation}. The outcome of the verification is encoded in a binary safety verdict \( v_t \in \{0, 1\} \), where \( v_t = 1 \) indicates that the trajectory meets all verification requirements.

\begin{figure}[t]
    \centering
    \includegraphics[width=0.55\linewidth]{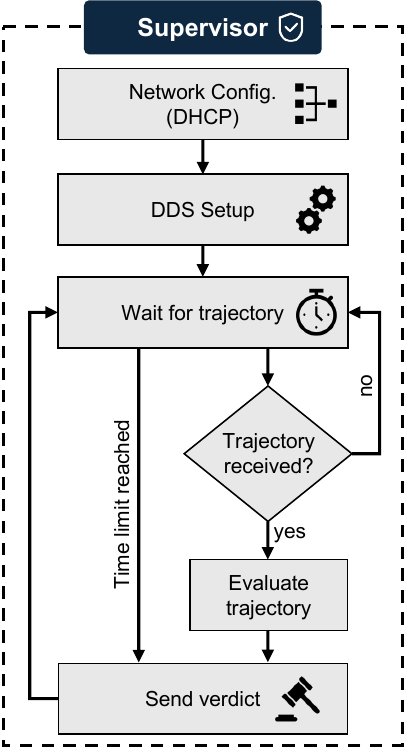}
    \caption{Execution loop of the Supervisor. If no trajectory is received within the time limit \( T_{\text{max}} \), the module issues a negative verdict. Otherwise, it evaluates the trajectory and publishes a binary result.}
    \label{fig:pipeline}
\end{figure}

In both cases, whether a trajectory is evaluated or a timeout occurs, the Supervisor publishes the current verdict for downstream use. This information can be consumed by safety controllers, vehicle actuation modules, or fallback mechanisms to initiate appropriate system responses. \autoref{fig:pipeline} illustrates the overall execution loop.

\subsection{Trajectory Evaluation}
\label{sec:evaluation}

The Supervisor verifies each received trajectory \( \xi = \{\mathbf{x}_0, \dots, \mathbf{x}_N\} \) against a defined set of safety and feasibility criteria. The evaluation is modular and can be adapted depending on the available interfaces and application requirements. In the current implementation, the verification includes kinematic feasibility checks and lightweight plausibility metrics~\cite{Trauth2024}. Each trajectory point \( \mathbf{x}_i\) (Equation~\ref{eq:x_i}) must satisfy basic kinematic constraints. These include acceleration bounds, curvature limits, curvature rate, and yaw rate constraints. The allowed acceleration is determined by a velocity-dependent envelope
\begin{equation}
a_{\text{permissible}}(v) =
\begin{cases}
a_{\text{max}} \cdot \frac{v_{\text{switch}}}{v}, & \text{if } v > v_{\text{switch}}, \\
a_{\text{max}}, & \text{otherwise}.
\end{cases}
\end{equation}
The trajectory is deemed kinematically feasible if its acceleration \( a(t) \) satisfies
\begin{equation}
-a_{\text{max}} \leq a(t) \leq a_{\text{permissible}}(t), \quad \forall t \in [t_0, t_f].
\end{equation}
The curvature \( \kappa(t) \) must remain below a limit based on the maximum steering angle \( \delta_\text{max} \) and the wheelbase \( L \)
\begin{equation}
|\kappa(t)| \leq \frac{\tan(\delta_\text{max})}{L}, \quad \forall t \in [t_0, t_f].
\end{equation}
Additional constraints apply to the rate of curvature change \( \dot{\kappa}(t) \) and yaw rate \( \dot{\psi}(t) \), where
\begin{align}
|\dot{\kappa}(t)| &\leq \dot{\kappa}_\text{max}, \quad \forall t \in [t_0, t_f], \\
|\dot{\psi}(t)| &\leq \kappa_\text{max} \cdot v(t), \quad \forall t \in [t_0, t_f].
\end{align}

Beyond feasibility, the Supervisor computes analytical plausibility metrics to assess geometric consistency and environmental context. These metrics include the lateral deviation from a given reference path \( \Gamma \)
\begin{equation}
J_{\text{ref}}(\xi) = \sum_{i=0}^{N} \left( d_i^\perp \right)^2,
\end{equation}
where \( d_i^\perp \) denotes the lateral distance from the trajectory point \( \mathbf{x}_i \) to the closest point on \( \Gamma \). Another example is the velocity offset cost, which penalizes deviation from a desired cruise speed \( v_{\text{desired}} \)
\begin{equation}
J_{\text{vel}}(\xi) = \sum_{i=0}^{N} \left| v_i - v_{\text{desired}} \right|^p,
\end{equation}
with \( p \in \{1,2\} \) indicating the norm used for penalization. Furthermore, comfort-related criteria are captured through acceleration-based costs. These lateral and longitudinal accelerations are evaluated via
\begin{equation}
J_{\text{lat}}(\xi) = \int_{t_0}^{t_f} a_{\text{lat}}(t)^2 \, \mathrm{d}t, \,
J_{\text{lon}}(\xi) = \int_{t_0}^{t_f} a_{\text{lon}}(t)^2 \, \mathrm{d}t.
\end{equation}
The presented cost functions are intended as representative examples, while
the framework allows additional criteria to be integrated as needed. In future iterations, it is planned to incorporate obstacle-aware metrics such as
\begin{equation}
J_{\text{obs}}(\xi) = \sum_{i=0}^{N} \sum_{j=1}^{M} \frac{1}{\text{dist}(\mathbf{x}_i, o_t^j)^2 + \epsilon},
\end{equation}
where \( \text{dist}(\cdot, \cdot) \) is the Euclidean distance to the predicted object \( o_t^j \), and \( \epsilon \) is a regularization term. 
At the current stage, however, no obstacle-related cost terms are computed.

As in other cost-based planning frameworks~\cite{Trauth2024}, the Supervisor combines the individual cost terms into a single scalar value. The total cost for a given trajectory \( \xi \) is computed as
\begin{equation}
J_{\text{sum}}(\xi \,|\, f_\xi) = \sum_{i=1}^{n} \omega_i \cdot J_i(\xi),
\end{equation}
where \( \omega_i \) is the weighting factor for the \( i \)-th cost component. This structure allows prioritization of different trajectory qualities. The total cost is evaluated alongside the feasibility checks and can be used to rank multiple trajectory candidates if required. Due to limited computational resources on the embedded hardware, no full drivability or reachability-based collision check is currently implemented. Instead, a lightweight fallback strategy is foreseen, which will verify spatial compatibility in a simplified manner using prefiltered bounding-box overlaps at discrete time steps. This mechanism is part of ongoing development.

If all mandatory criteria are fulfilled, the trajectory is marked as valid and a positive verdict is issued. Otherwise, the Supervisor rejects the trajectory and triggers the corresponding fallback mechanism.

%%%%%%%%%%%%%%%%%%%%%%%%%%%%%%%%%%%%%%%%%%%%%%%%%%%%%%%%%
%%% Results
%%%%%%%%%%%%%%%%%%%%%%%%%%%%%%%%%%%%%%%%%%%%%%%%%%%%%%%%%
\section{Results \& Analysis}
\label{sec:results}

This section presents the performance evaluation of the proposed Supervisor, with a particular focus on its execution time. The goal is to determine whether the Supervisor logic can be executed reliably on embedded platforms within bounded cycle times and without violating real-time requirements. We compare execution latencies across different hardware platforms and analyze timing consistency, outliers, and communication behavior. All measurements were obtained under controlled input conditions with prerecorded trajectories.

\subsection{Evaluation Setup}
\label{sec:evaluation-setup}

The system was deployed on two embedded platforms based on the NXP S32Z2 real-time processor: the official NXP Evaluation Board and the ARM Cortex-R Automotive Development System. Both platforms use a single-core ARM Cortex-R52 running at \SI{800}{\mega\hertz} under Zephyr RTOS (v3.5). Due to their identical processor configuration, preliminary measurements revealed only negligible differences in runtime behavior. As a result, the evaluation in this paper focuses on the ARM platform, while noting that equivalent performance can be expected on the NXP board. For reference, all experiments were also repeated on a high-performance desktop system featuring an AMD Ryzen 7840HS (\SI{5.1}{\giga\hertz}) running Ubuntu 22.04. The Supervisor was executed as a standalone task on each system. To evaluate real-time behavior, we implemented a lightweight timing mechanism based on CPU cycle counters, using built-in system APIs to capture start and end timestamps. The timing resolution corresponds to the processor clock frequency and is converted into nanoseconds based on the known system tick rate. All platforms received identical prerecorded trajectories over a DDS-based interface. The Supervisor was triggered by message arrival and executed exactly once per trajectory.

\subsection{Runtime Performance and Timing Analysis}
\label{sec:timingresults}

To evaluate timing behavior, we report the average runtime $T_{\text{avg}}$, the worst-case runtime $T_{\text{max}}$, and the jitter $J$, defined as
\begin{align}
T_{\text{avg}} &= \frac{1}{n} \sum_{i=1}^{n} T_i, \label{eq:avg}\\
T_{\text{max}} &= \max(T_i), \label{eq:max} \vphantom{\sum_{i=1}^{n} T_i}\\
J &= \max \left( \left| T_i - T_{\text{avg}} \right| \right) \vphantom{\sum T_i}, \label{eq:jitter}
\end{align}
where \( T_i \) denotes the runtime of the \( i \)-th trajectory evaluation and \( n \) is the number of samples.

\begin{figure}[!ht]
    \centering
    \input{figures/latency-arm-vs-hpc}
    \vspace{-0.4cm}
    \caption{Measured runtime per trajectory evaluation on the ARM Development System and HPC platform. While the HPC system achieves lower average runtimes, its variability is significantly higher. The ARM platform exhibits more stable and predictable execution behavior.}
    \label{fig:runtime_arm_vs_hpc}%
\end{figure}
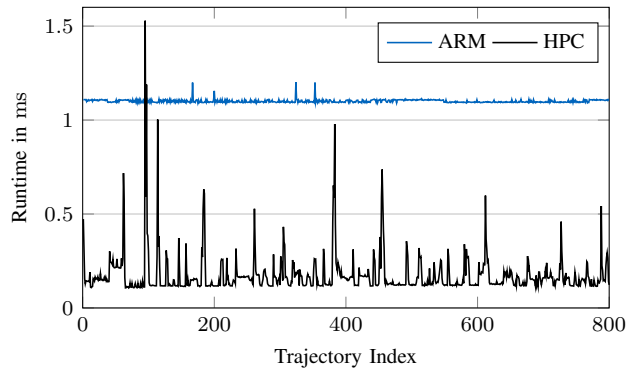

Figure~\ref{fig:runtime_arm_vs_hpc} shows a comparison of 800 sequential evaluations on the ARM and HPC systems. As expected, the HPC system achieves lower average runtimes but exhibits considerably higher temporal variability. In contrast, the ARM system demonstrates much more consistent performance with visibly reduced jitter.

This observation is quantitatively supported by the results in Table~\ref{tab:benchmark_ms_comparison}. While the ARM system processes each trajectory in approximately \SI{1.1}{\milli\second} on average, the HPC achieves a runtime of only \SI{0.17}{\milli\second}. However, the jitter on the HPC system exceeds \SI{778}{\percent} of its average runtime, compared to just \SI{9.3}{\percent} on the ARM platform. Despite the higher absolute latency, the real-time behavior of the ARM system is significantly more predictable and thus more suitable for safety-relevant deployment.

\begin{table}[!ht]
\centering
\caption{Timing metrics comparing the ARM System and HPC platform.}
\label{tab:benchmark_ms_comparison}
\begin{tabular}{lrrr}
\toprule
\textbf{Metric} & \textbf{ARM} & \textbf{HPC} & \textbf{Relative Diff. (\%)} \\
\midrule
Minimum runtime (\si{\milli\second})    & 1.089 & 0.107 & -90.17 \\
Maximum runtime (\si{\milli\second})    & 1.202 & 1.529 & 27.20  \\
Average runtime (\si{\milli\second})    & 1.100 & 0.174 & -84.18 \\
Jitter (\si{\milli\second})             & 0.102 & 1.355 & 1228.43 \\
Jitter (\si{\percent})                  & 9.27 & 778.74 & 8,300 \\
\bottomrule
\end{tabular}
\end{table}

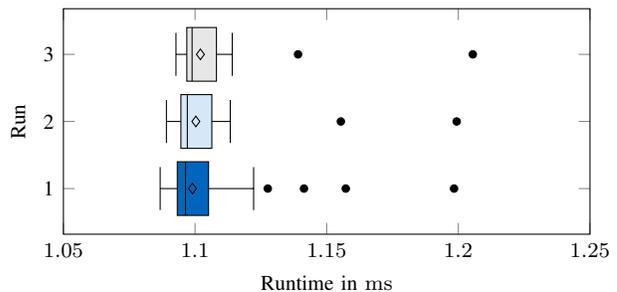
\begin{figure}[!ht]
    \centering
    \input{figures/runtime_boxplot}
    \vspace{-0.4cm}
    \caption{Boxplot of three evaluation runs with 10,000 trajectories each on the ARM Development System. The results show tight runtime clustering with few outliers and low overall jitter.}
    \label{fig:runtime_boxplot}%
\end{figure}

To investigate runtime stability further, we performed three evaluation runs with 10,000 trajectories each on the ARM System. Figure~\ref{fig:runtime_boxplot} shows the resulting boxplots. Across all runs, the latencies remain tightly clustered, with only a few outliers. These deviations appear randomly and do not correlate with specific trajectory indices, suggesting that the underlying cause is unrelated to input structure and more likely rooted in occasional system-level effects such as memory contention or interrupt handling.

Initial experiments also revealed occasional increases in round-trip latency, particularly under high middleware load. While the Supervisor itself remains temporally stable, DDS message delivery and deserialization introduce variability that must be addressed in future work to ensure reliable integration into online systems.

\subsection{Functional Verification Results}
\label{sec:verificationresults}

In addition to timing metrics, we conducted a qualitative inspection of the Supervisor's behavior by logging selected trajectory inputs and their corresponding feasibility verdicts. This complements the performance analysis by illustrating the practical operation of the current implementation. Since the presented system represents an early prototype, the evaluation results serve as a first benchmark and basis for further refinement.

\begin{figure*}[ht!]
    \centering
    \input{figures/trajectory_colors}
    \subfloat[][\scriptsize{Evaluation results in a curvilinear coordinate system.}]{\label{fig:trajectories_s_d_cost} \input{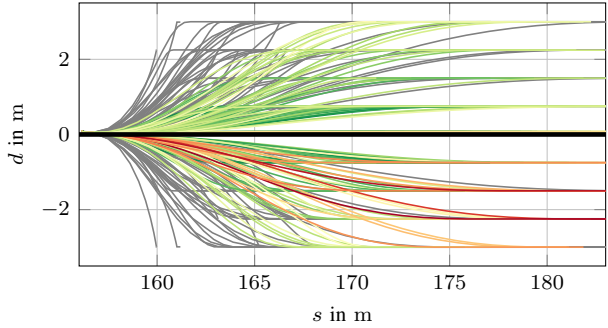}} \hspace{0.5cm}
    \subfloat[][\scriptsize{Evaluation results in a Cartesian coordinate system.}]{\label{fig:trajectories_x_y_cost} \input{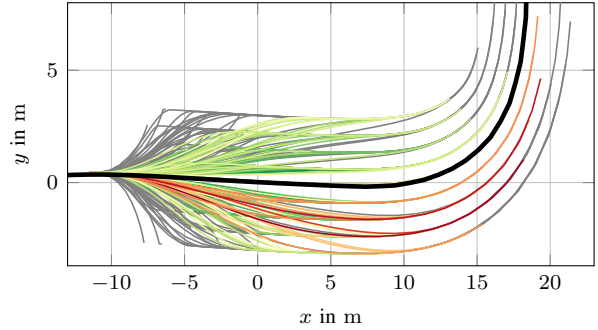}} \\
    \caption{Evaluation of trajectory candidates by the Supervisor along a given reference path (\protect\refpath). Feasibility results are shown in gray for invalid trajectories. Valid trajectories are color-coded based on their total cost, ranging from green (low cost) to red (high cost).}
    \label{fig:qualitative_evaluation}
\end{figure*}

Figure~\ref{fig:qualitative_evaluation} illustrates an exemplary test case, where the Supervisor evaluates a set of trajectories. Figure~\ref{fig:trajectories_s_d_cost} shows the trajectories in curvilinear coordinates along the reference path, while Figure~\ref{fig:trajectories_x_y_cost} displays the corresponding representation in Cartesian space. Trajectories that fail the feasibility check are shown in gray. These typically exceed dynamic constraints such as maximum curvature or acceleration and are therefore marked as infeasible. All other trajectories are colored according to their total cost, ranging from green (low cost) to red (high cost).

The visualization provides an interpretable representation of the cost and feasibility evaluation, enabling validation of the implemented decision logic. For this illustrative case, all generated trajectories were evaluated. In a real system deployment, however, only a subset of candidates, typically the optimal trajectory selected by the motion planner, would be sent to the Supervisor for verification. The visualization also highlights the influence of the cost structure: trajectories with strong lateral deviation from the reference path or excessive positive acceleration are penalized accordingly. Similarly, deviations from a target cruise speed of \SI{7}{\meter\per\second} contribute to the total cost. No parameter tuning was performed in this example. All cost weights were selected manually, and the target speed was fixed for illustrative purposes only.

Since the current prototype does not operate in an online loop, trajectory messages were provided from a prerecorded dataset. Synthetic timing gaps were inserted between messages to simulate delays. The Supervisor correctly detected violations and logged warnings, but no reactive behavior was triggered at this stage.

%%%%%%%%%%%%%%%%%%%%%%%%%%%%%%%%%%%%%%%%%%%%%%%%%%%%%%%%%
%%% Discussion
%%%%%%%%%%%%%%%%%%%%%%%%%%%%%%%%%%%%%%%%%%%%%%%%%%%%%%%%%
\section{Discussion}
\label{sec:discussion}

The presented safeguarding system constitutes a functional prototype capable of verifying motion planning trajectories in real-time. The current implementation demonstrates that trajectory evaluation based on kinematic constraints and analytical cost functions can be performed with low execution latency and minimal jitter on embedded hardware. This confirms the basic feasibility of the proposed approach under the outlined architectural conditions.

However, several limitations remain. The system is currently not integrated into an online planning loop and processes precomputed trajectories. While timing violations can be detected via a time safeguard mechanism, no fallback behavior is currently implemented.

Preliminary results indicate that DDS-based communication is viable for data transmission, but observations show that message processing latency can fluctuate, resulting in increased round-trip times. These delays must be systematically analyzed and addressed to ensure reliable operation in online scenarios. Optimizing the data interface and reducing middleware-induced variance will be key to deploying the Supervisor in a fully integrated AV stack.

Overall, the current results are promising and validate the core concept of modular runtime verification. Further integration and robustness testing are required to move from a prototypical implementation to a deployable AV safeguarding system.

%%%%%%%%%%%%%%%%%%%%%%%%%%%%%%%%%%%%%%%%%%%%%%%%%%%%%%%%%
%%% Conclusion
%%%%%%%%%%%%%%%%%%%%%%%%%%%%%%%%%%%%%%%%%%%%%%%%%%%%%%%%%
\section{Conclusion \& Outlook}
\label{sec:conclusion}

This paper presented a modular safeguarding system for trajectory planning in AVs, designed to operate independently on embedded real-time hardware. The proposed Supervisor evaluates incoming trajectories with respect to feasibility constraints and plausibility costs, while a conceptual time safeguard monitors the temporal consistency of planning outputs. The system has been prototypically implemented on two automotive-grade platforms based on the ARM Cortex-R52 architecture and benchmarked against a high-performance reference system.

Initial results confirm that the Supervisor can execute trajectory evaluations within tight timing bounds and with minimal jitter. Qualitative inspection of feasibility verdicts further illustrates that the implemented cost functions penalize implausible or unsafe behaviors as expected. Communication with the main planning module is handled via DDS, enabling modular integration into existing software stacks.

Future work will extend the Supervisor's ability to actively respond to detected safety violations. A dedicated fallback module is foreseen, based on a simplified planning component capable of generating minimum risk maneuvers under limited information. This module will generate backup trajectories using risk-aware metrics and ensure a valid action is available even in degraded planning conditions.

In addition, the Supervisor will be integrated into an online planning loop, first in simulation and later on a real test vehicle. This will allow end-to-end evaluation of the full safeguarding pipeline under realistic runtime conditions and provide a foundation for certification-oriented development in safety-critical AV applications.

%%%%%%%%%%%%%%%%%%%%%%%%%%%%%%%%%%%%%%%%%%%%%%%%%%%%%%%%%
%%% Acknowledgement
%%%%%%%%%%%%%%%%%%%%%%%%%%%%%%%%%%%%%%%%%%%%%%%%%%%%%%%%%

%%%%%%%%%%%%%%%%%%%%%%%%%%%%%%%%%%%%%%%%%%%%%%%%%%%%%%%%%%%%%%%%%%%%%%%%%%%%%%%%

\bibliographystyle{IEEEtran}
\bibliography{literature}
\end{document}

%% file: figures/latency-arm-vs-hpc.tex
\begin{tikzpicture}[font=\footnotesize]
\begin{axis}[
width=7cm,
height=4.0cm,
scale only axis,
scaled ticks=false,
scaled ticks=false,
xlabel = {Trajectory Index},
ylabel = {Runtime in \si{\milli\second}},
ymin = 0, ymax = 1.6,
xmin = 0, xmax = 800,
ymajorgrids=true,
major grid style={line width=.2pt,draw=gray!50},
legend style={
    at={(0.77, 0.95)}, 
    anchor=north,
    legend columns=2,
    cells={anchor=center}
}
]
\addplot[Blue, semithick] table{
1	1.10925
2	1.108375
3	1.109125
4	1.10875
5	1.10925
6	1.098375
7	1.104875
8	1.10275
9	1.107625
10	1.103
11	1.105125
12	1.1065
13	1.105375
14	1.10925
15	1.099125
16	1.1065
17	1.106125
18	1.11
19	1.106625
20	1.104875
21	1.10675
22	1.1035
23	1.10675
24	1.102375
25	1.104125
26	1.10925
27	1.107875
28	1.108625
29	1.110125
30	1.107
31	1.10675
32	1.10675
33	1.105
34	1.108
35	1.10375
36	1.10675
37	1.108625
38	1.094625
39	1.093875
40	1.095875
41	1.094875
42	1.096625
43	1.093875
44	1.0935
45	1.095625
46	1.09675
47	1.09575
48	1.097875
49	1.097375
50	1.095375
51	1.107625
52	1.0935
53	1.098375
54	1.104625
55	1.09625
56	1.1075
57	1.105125
58	1.107375
59	1.106125
60	1.106
61	1.10475
62	1.10675
63	1.106375
64	1.1075
65	1.1085
66	1.1055
67	1.105875
68	1.108125
69	1.1055
70	1.108
71	1.097
72	1.09875
73	1.105
74	1.097
75	1.094625
76	1.10925
77	1.09525
78	1.0945
79	1.097125
80	1.10725
81	1.094875
82	1.10875
83	1.090875
84	1.105
85	1.09425
86	1.108625
87	1.09425
88	1.1065
89	1.092375
90	1.0935
91	1.0925
92	1.105125
93	1.09575
94	1.1095
95	1.0955
96	1.104875
97	1.09525
98	1.1065
99	1.092125
100	1.093125
101	1.105875
102	1.093875
103	1.10625
104	1.09275
105	1.108
106	1.09625
107	1.0955
108	1.0965
109	1.094875
110	1.094625
111	1.094
112	1.094375
113	1.09125
114	1.094125
115	1.094375
116	1.1115
117	1.096375
118	1.108
119	1.106875
120	1.093375
121	1.09825
122	1.1065
123	1.095375
124	1.107125
125	1.10375
126	1.1105
127	1.105875
128	1.092875
129	1.105125
130	1.094375
131	1.096
132	1.109125
133	1.09425
134	1.10775
135	1.094625
136	1.0965
137	1.1045
138	1.10675
139	1.09525
140	1.1045
141	1.0935
142	1.104875
143	1.093875
144	1.093875
145	1.10575
146	1.0945
147	1.106125
148	1.097875
149	1.095
150	1.10825
151	1.09925
152	1.105125
153	1.095125
154	1.096125
155	1.105125
156	1.09375
157	1.089875
158	1.106875
159	1.10275
160	1.09625
161	1.097625
162	1.095625
163	1.0915
164	1.10675
165	1.093625
166	1.0945
167	1.1995
168	1.105625
169	1.095375
170	1.0975
171	1.093625
172	1.104875
173	1.108125
174	1.1065
175	1.090875
176	1.105625
177	1.09275
178	1.097625
179	1.096125
180	1.109375
181	1.093375
182	1.093375
183	1.092625
184	1.09325
185	1.095
186	1.09425
187	1.10775
188	1.09475
189	1.10725
190	1.096
191	1.0975
192	1.093125
193	1.095
194	1.0975
195	1.0965
196	1.096
197	1.09425
198	1.095125
199	1.110625
200	1.155375
201	1.09375
202	1.10525
203	1.105375
204	1.09475
205	1.107125
206	1.096125
207	1.0945
208	1.09525
209	1.108
210	1.094125
211	1.0955
212	1.109375
213	1.092625
214	1.096
215	1.09475
216	1.106625
217	1.091625
218	1.106375
219	1.09375
220	1.10775
221	1.096625
222	1.094625
223	1.10475
224	1.0975
225	1.0985
226	1.097875
227	1.106875
228	1.09425
229	1.096375
230	1.092875
231	1.0955
232	1.09325
233	1.095625
234	1.10175
235	1.096125
236	1.103125
237	1.10475
238	1.095375
239	1.094375
240	1.09375
241	1.095
242	1.105875
243	1.0925
244	1.09175
245	1.094
246	1.107875
247	1.09625
248	1.104375
249	1.0955
250	1.094625
251	1.093875
252	1.096125
253	1.09525
254	1.094375
255	1.107875
256	1.0955
257	1.103
258	1.09775
259	1.0945
260	1.093625
261	1.104
262	1.0965
263	1.10725
264	1.089125
265	1.096125
266	1.095875
267	1.1065
268	1.095125
269	1.093875
270	1.095
271	1.099
272	1.098875
273	1.09775
274	1.0975
275	1.09275
276	1.096625
277	1.1065
278	1.094375
279	1.094375
280	1.106375
281	1.108
282	1.109375
283	1.097125
284	1.10525
285	1.095625
286	1.0965
287	1.098125
288	1.10625
289	1.096
290	1.09725
291	1.097375
292	1.097
293	1.095375
294	1.093
295	1.093875
296	1.09575
297	1.095625
298	1.109375
299	1.09575
300	1.100125
301	1.095
302	1.095625
303	1.101625
304	1.09875
305	1.095875
306	1.106625
307	1.094125
308	1.092125
309	1.097125
310	1.096
311	1.106875
312	1.096375
313	1.096
314	1.0965
315	1.091375
316	1.095375
317	1.09525
318	1.10125
319	1.0935
320	1.094875
321	1.0945
322	1.09275
323	1.094375
324	1.202
325	1.110625
326	1.108375
327	1.093
328	1.10625
329	1.098875
330	1.09525
331	1.096625
332	1.10575
333	1.10575
334	1.094375
335	1.093875
336	1.095875
337	1.09275
338	1.096375
339	1.105625
340	1.093875
341	1.10375
342	1.106375
343	1.096
344	1.10375
345	1.095
346	1.096
347	1.098625
348	1.0965
349	1.09575
350	1.09375
351	1.1005
352	1.095
353	1.199375
354	1.0935
355	1.1105
356	1.104875
357	1.092875
358	1.10425
359	1.09625
360	1.10625
361	1.099375
362	1.09825
363	1.09525
364	1.108125
365	1.093875
366	1.105
367	1.09725
368	1.09475
369	1.097375
370	1.109625
371	1.107625
372	1.093375
373	1.097
374	1.092875
375	1.108
376	1.107875
377	1.10325
378	1.097625
379	1.09325
380	1.096
381	1.09525
382	1.10675
383	1.095
384	1.093875
385	1.09525
386	1.09375
387	1.095
388	1.0945
389	1.091875
390	1.097125
391	1.103875
392	1.096
393	1.093875
394	1.09625
395	1.09725
396	1.09525
397	1.09375
398	1.09575
399	1.09425
400	1.104625
401	1.105375
402	1.107125
403	1.106125
404	1.0945
405	1.1065
406	1.0945
407	1.105375
408	1.104125
409	1.101
410	1.097125
411	1.092375
412	1.095
413	1.097375
414	1.094
415	1.095625
416	1.10425
417	1.10525
418	1.105875
419	1.093375
420	1.09775
421	1.09675
422	1.094
423	1.095875
424	1.09175
425	1.10925
426	1.10675
427	1.0955
428	1.09925
429	1.092125
430	1.096875
431	1.105125
432	1.092625
433	1.097375
434	1.095625
435	1.09925
436	1.0945
437	1.094125
438	1.105875
439	1.106375
440	1.107375
441	1.1085
442	1.11
443	1.111375
444	1.107375
445	1.107875
446	1.094
447	1.106125
448	1.091875
449	1.104625
450	1.10625
451	1.108
452	1.1105
453	1.09375
454	1.094375
455	1.1075
456	1.106
457	1.109125
458	1.108375
459	1.107
460	1.11075
461	1.09875
462	1.09475
463	1.094625
464	1.103625
465	1.106625
466	1.105375
467	1.094125
468	1.093375
469	1.106625
470	1.107
471	1.096625
472	1.1095
473	1.106375
474	1.109625
475	1.1085
476	1.1055
477	1.091375
478	1.1075
479	1.10725
480	1.108
481	1.10775
482	1.108875
483	1.1085
484	1.107
485	1.108875
486	1.108375
487	1.10575
488	1.1025
489	1.10775
490	1.1125
491	1.10625
492	1.1095
493	1.10725
494	1.105375
495	1.10775
496	1.10775
497	1.107875
498	1.10775
499	1.105
500	1.104125
501	1.108875
502	1.10675
503	1.1095
504	1.108125
505	1.108
506	1.1045
507	1.1045
508	1.109625
509	1.106
510	1.1075
511	1.10725
512	1.109125
513	1.11
514	1.1085
515	1.113375
516	1.105875
517	1.10625
518	1.104625
519	1.107
520	1.10875
521	1.107
522	1.10775
523	1.106375
524	1.106125
525	1.110375
526	1.105375
527	1.108
528	1.1065
529	1.110125
530	1.10825
531	1.10675
532	1.10675
533	1.10925
534	1.108875
535	1.106625
536	1.108
537	1.104875
538	1.106375
539	1.107375
540	1.105875
541	1.106375
542	1.11025
543	1.105875
544	1.1045
545	1.104375
546	1.10575
547	1.10675
548	1.10725
549	1.09475
550	1.094875
551	1.09475
552	1.107
553	1.094
554	1.094625
555	1.092375
556	1.092625
557	1.09375
558	1.09725
559	1.09225
560	1.0925
561	1.096875
562	1.09625
563	1.095875
564	1.094625
565	1.095875
566	1.099625
567	1.0915
568	1.093125
569	1.099625
570	1.095625
571	1.094375
572	1.09525
573	1.094375
574	1.095125
575	1.094125
576	1.095875
577	1.09525
578	1.093
579	1.096
580	1.095125
581	1.093375
582	1.09375
583	1.091125
584	1.10475
585	1.0945
586	1.095625
587	1.09625
588	1.09625
589	1.0945
590	1.095
591	1.09375
592	1.095
593	1.09475
594	1.09475
595	1.09725
596	1.0925
597	1.095125
598	1.09575
599	1.108
600	1.09425
601	1.09475
602	1.096375
603	1.097
604	1.092875
605	1.092125
606	1.093875
607	1.094
608	1.093125
609	1.096375
610	1.094875
611	1.095
612	1.097375
613	1.096625
614	1.09475
615	1.098125
616	1.09475
617	1.09725
618	1.09525
619	1.0965
620	1.094
621	1.092875
622	1.094125
623	1.09375
624	1.098375
625	1.105875
626	1.094125
627	1.093875
628	1.096125
629	1.09525
630	1.093125
631	1.094125
632	1.09825
633	1.096
634	1.097625
635	1.093625
636	1.097875
637	1.09475
638	1.09325
639	1.0955
640	1.093375
641	1.09525
642	1.09725
643	1.094125
644	1.09775
645	1.09425
646	1.095375
647	1.095125
648	1.096
649	1.094125
650	1.09625
651	1.094375
652	1.108875
653	1.095
654	1.092875
655	1.095625
656	1.0965
657	1.0945
658	1.105625
659	1.09625
660	1.093
661	1.0935
662	1.094375
663	1.099
664	1.095
665	1.098625
666	1.11
667	1.09625
668	1.09475
669	1.09725
670	1.096375
671	1.095
672	1.0945
673	1.09475
674	1.09875
675	1.097375
676	1.097875
677	1.0935
678	1.109625
679	1.093625
680	1.095875
681	1.10525
682	1.095625
683	1.09625
684	1.0925
685	1.108
686	1.09425
687	1.095625
688	1.09225
689	1.09675
690	1.09425
691	1.097
692	1.096375
693	1.0955
694	1.09475
695	1.098125
696	1.093
697	1.0965
698	1.095
699	1.095625
700	1.09325
701	1.093875
702	1.095625
703	1.09425
704	1.09825
705	1.094625
706	1.094125
707	1.09425
708	1.10925
709	1.09525
710	1.10725
711	1.097125
712	1.106
713	1.104375
714	1.096875
715	1.0965
716	1.094375
717	1.107375
718	1.094875
719	1.0955
720	1.09375
721	1.095
722	1.09625
723	1.094125
724	1.092875
725	1.096625
726	1.0935
727	1.09725
728	1.094375
729	1.09425
730	1.09575
731	1.093125
732	1.09575
733	1.097
734	1.095
735	1.0945
736	1.0945
737	1.096125
738	1.10625
739	1.0945
740	1.093375
741	1.100375
742	1.0935
743	1.09725
744	1.10775
745	1.092625
746	1.108125
747	1.096875
748	1.094375
749	1.1045
750	1.098375
751	1.09375
752	1.097875
753	1.095125
754	1.092875
755	1.09725
756	1.09375
757	1.095625
758	1.096625
759	1.094875
760	1.09475
761	1.108375
762	1.09875
763	1.105875
764	1.095625
765	1.107625
766	1.1065
767	1.091375
768	1.09375
769	1.093625
770	1.109875
771	1.107875
772	1.107875
773	1.107875
774	1.108375
775	1.105375
776	1.108625
777	1.1095
778	1.105875
779	1.106375
780	1.105125
781	1.110125
782	1.106375
783	1.1065
784	1.10675
785	1.10725
786	1.109
787	1.105625
788	1.108
789	1.10775
790	1.106625
791	1.106875
792	1.106125
793	1.10475
794	1.104125
795	1.10625
796	1.106375
797	1.10875
798	1.104125
799	1.10925
800	1.108375
};
\addlegendentry{ARM}
\addplot[Black, semithick] table {
1	0.473114
2	0.313465
3	0.156402
4	0.128718
5	0.143067
6	0.144419
7	0.141954
8	0.141463
9	0.142065
10	0.14009
11	0.191079
12	0.110914
13	0.108149
14	0.111996
15	0.114431
16	0.156643
17	0.151643
18	0.143498
19	0.164738
20	0.155741
21	0.157103
22	0.153456
23	0.16546
24	0.146863
25	0.151663
26	0.149399
27	0.182693
28	0.151733
29	0.149338
30	0.151322
31	0.145752
32	0.147324
33	0.162995
34	0.153046
35	0.163566
36	0.143517
37	0.149869
38	0.139559
39	0.139659
40	0.140622
41	0.301422
42	0.246576
43	0.242018
44	0.241486
45	0.232248
46	0.259862
47	0.219814
48	0.214104
49	0.214034
50	0.215616
51	0.216729
52	0.261225
53	0.219725
54	0.215456
55	0.213642
56	0.215196
57	0.211909
58	0.213332
59	0.26407
60	0.22284
61	0.261214
62	0.717576
63	0.568377
64	0.162103
65	0.106947
66	0.107638
67	0.114121
68	0.112517
69	0.113018
70	0.106505
71	0.109682
72	0.133538
73	0.112156
74	0.110964
75	0.131714
76	0.112447
77	0.108159
78	0.107698
79	0.112717
80	0.130081
81	0.115804
82	0.10851
83	0.107768
84	0.131144
85	0.110654
86	0.109892
87	0.107819
88	0.130502
89	0.111375
90	0.108419
91	0.142335
92	0.117777
93	0.111295
94	0.111816
95	1.528822
96	0.594077
97	1.188865
98	0.397006
99	0.383791
100	0.302965
101	0.134389
102	0.121214
103	0.11902
104	0.119291
105	0.118508
106	0.121144
107	0.120582
108	0.119621
109	0.118579
110	0.117857
111	0.117917
112	0.118028
113	0.272216
114	1.003027
115	0.328184
116	0.38312
117	0.218702
118	0.118348
119	0.117657
120	0.118409
121	0.117808
122	0.117116
123	0.117336
124	0.119
125	0.117166
126	0.117707
127	0.306441
128	0.26428
129	0.261064
130	0.118569
131	0.120362
132	0.126785
133	0.118008
134	0.119721
135	0.116796
136	0.117177
137	0.118048
138	0.149759
139	0.148798
140	0.120462
141	0.117667
142	0.117316
143	0.118419
144	0.11916
145	0.115984
146	0.372309
147	0.226397
148	0.118509
149	0.115843
150	0.117096
151	0.116184
152	0.115913
153	0.114561
154	0.115143
155	0.115152
156	0.114742
157	0.342852
158	0.184075
159	0.118519
160	0.148126
161	0.118529
162	0.118128
163	0.119961
164	0.149028
165	0.147264
166	0.149329
167	0.15014
168	0.147314
169	0.156051
170	0.152084
171	0.153467
172	0.152334
173	0.15534
174	0.148577
175	0.117617
176	0.11899
177	0.117958
178	0.11394
179	0.117727
180	0.119742
181	0.321811
182	0.270893
183	0.526897
184	0.631419
185	0.609086
186	0.291502
187	0.117247
188	0.116505
189	0.117056
190	0.116024
191	0.118709
192	0.11894
193	0.11929
194	0.116365
195	0.129681
196	0.140311
197	0.120723
198	0.11898
199	0.116785
200	0.117166
201	0.117066
202	0.117257
203	0.117366
204	0.116846
205	0.117366
206	0.116936
207	0.146022
208	0.121184
209	0.120192
210	0.247047
211	0.259411
212	0.257026
213	0.258178
214	0.119932
215	0.118088
216	0.11885
217	0.124761
218	0.124611
219	0.266455
220	0.157724
221	0.164808
222	0.127006
223	0.12941
224	0.118159
225	0.117406
226	0.118048
227	0.117096
228	0.132877
229	0.148246
230	0.144369
231	0.14472
232	0.130942
233	0.315569
234	0.210918
235	0.167794
236	0.16565
237	0.164197
238	0.16617
239	0.162183
240	0.162233
241	0.166702
242	0.164838
243	0.16526
244	0.163816
245	0.168986
246	0.163135
247	0.163195
248	0.165639
249	0.167082
250	0.168045
251	0.166431
252	0.169067
253	0.154759
254	0.147605
255	0.145661
256	0.148196
257	0.156021
258	0.131103
259	0.123228
260	0.267937
261	0.52798
262	0.337992
263	0.196971
264	0.178204
265	0.174928
266	0.17613
267	0.178305
268	0.175549
269	0.12981
270	0.118869
271	0.118899
272	0.117446
273	0.136083
274	0.152465
275	0.192942
276	0.203894
277	0.206138
278	0.173075
279	0.16576
280	0.158856
281	0.128859
282	0.123218
283	0.118208
284	0.118469
285	0.116905
286	0.116105
287	0.116975
288	0.117216
289	0.116906
290	0.285301
291	0.166752
292	0.169687
293	0.162684
294	0.16031
295	0.131274
296	0.123739
297	0.148246
298	0.151332
299	0.163917
300	0.121906
301	0.27437
302	0.245324
303	0.216017
304	0.146242
305	0.431644
306	0.351769
307	0.33685
308	0.163896
309	0.167023
310	0.164137
311	0.15526
312	0.130152
313	0.120532
314	0.11886
315	0.118529
316	0.117096
317	0.140642
318	0.137165
319	0.255293
320	0.208823
321	0.242739
322	0.162473
323	0.184957
324	0.188434
325	0.179737
326	0.18617
327	0.197732
328	0.199576
329	0.119491
330	0.203964
331	0.179206
332	0.169317
333	0.168675
334	0.1295
335	0.11918
336	0.11917
337	0.118499
338	0.11915
339	0.120142
340	0.125402
341	0.166091
342	0.180218
343	0.168275
344	0.163996
345	0.164468
346	0.140862
347	0.117687
348	0.116816
349	0.117797
350	0.118809
351	0.206479
352	0.261645
353	0.258329
354	0.259832
355	0.179847
356	0.124771
357	0.145922
358	0.121855
359	0.120573
360	0.11884
361	0.117567
362	0.117878
363	0.11914
364	0.119531
365	0.117246
366	0.314618
367	0.270562
368	0.129881
369	0.12446
370	0.126084
371	0.123379
372	0.122567
373	0.121615
374	0.123388
375	0.142826
376	0.12437
377	0.118428
378	0.11947
379	0.125864
380	0.460209
381	0.652189
382	0.587715
383	0.978699
384	0.446192
385	0.239142
386	0.218162
387	0.198093
388	0.242138
389	0.226577
390	0.21217
391	0.174848
392	0.16586
393	0.194866
394	0.161753
395	0.162985
396	0.164428
397	0.166722
398	0.16039
399	0.159789
400	0.159297
401	0.151101
402	0.154118
403	0.145691
404	0.146754
405	0.152544
406	0.162514
407	0.164318
408	0.167183
409	0.164678
410	0.163435
411	0.311562
412	0.170429
413	0.120213
414	0.121515
415	0.120563
416	0.120343
417	0.120783
418	0.119821
419	0.11903
420	0.214765
421	0.173485
422	0.17086
423	0.171151
424	0.17121
425	0.168906
426	0.167924
427	0.166943
428	0.168565
429	0.17599
430	0.169166
431	0.16583
432	0.166712
433	0.167965
434	0.207671
435	0.173656
436	0.143547
437	0.117126
438	0.116986
439	0.117006
440	0.117457
441	0.117868
442	0.311231
443	0.269902
444	0.253329
445	0.124861
446	0.119842
447	0.120353
448	0.11904
449	0.122095
450	0.130321
451	0.280091
452	0.37786
453	0.210136
454	0.588035
455	0.737143
456	0.573207
457	0.412176
458	0.319617
459	0.143928
460	0.126915
461	0.147184
462	0.137906
463	0.12399
464	0.12433
465	0.127426
466	0.12923
467	0.130762
468	0.132836
469	0.125793
470	0.125342
471	0.141945
472	0.126324
473	0.121865
474	0.121645
475	0.121705
476	0.121094
477	0.121505
478	0.12446
479	0.124941
480	0.122196
481	0.131724
482	0.124501
483	0.121504
484	0.122958
485	0.121855
486	0.123889
487	0.123488
488	0.122066
489	0.122898
490	0.121865
491	0.124841
492	0.355125
493	0.32093
494	0.278928
495	0.150531
496	0.125322
497	0.123328
498	0.122206
499	0.120573
500	0.124401
501	0.121315
502	0.125702
503	0.122958
504	0.123789
505	0.120944
506	0.122467
507	0.135592
508	0.162574
509	0.165429
510	0.162104
511	0.319327
512	0.274751
513	0.268479
514	0.273027
515	0.189265
516	0.119711
517	0.119682
518	0.121214
519	0.117917
520	0.130141
521	0.120583
522	0.122126
523	0.120552
524	0.145441
525	0.124641
526	0.124511
527	0.210607
528	0.1294
529	0.122727
530	0.123328
531	0.133919
532	0.126775
533	0.152244
534	0.242558
535	0.192272
536	0.13999
537	0.130632
538	0.134199
539	0.139709
540	0.137014
541	0.157404
542	0.185208
543	0.218152
544	0.255414
545	0.251035
546	0.159287
547	0.12975
548	0.125352
549	0.121324
550	0.119451
551	0.122807
552	0.122847
553	0.122887
554	0.120814
555	0.314658
556	0.275483
557	0.18635
558	0.121755
559	0.120433
560	0.121484
561	0.122406
562	0.118098
563	0.121254
564	0.121224
565	0.119952
566	0.121845
567	0.12932
568	0.119942
569	0.143908
570	0.125091
571	0.123107
572	0.124561
573	0.121765
574	0.121785
575	0.120322
576	0.121555
577	0.130873
578	0.154288
579	0.131404
580	0.339315
581	0.273078
582	0.172553
583	0.314036
584	0.273779
585	0.268379
586	0.271524
587	0.211999
588	0.122126
589	0.122547
590	0.121064
591	0.120944
592	0.121535
593	0.122085
594	0.120493
595	0.120783
596	0.121054
597	0.127306
598	0.120683
599	0.121124
600	0.121865
601	0.120623
602	0.244432
603	0.183474
604	0.179296
605	0.177332
606	0.19173
607	0.188615
608	0.186891
609	0.234653
610	0.229664
611	0.229884
612	0.598185
613	0.366779
614	0.313545
615	0.246365
616	0.253439
617	0.207741
618	0.133969
619	0.127065
620	0.12957
621	0.130472
622	0.127106
623	0.125743
624	0.127837
625	0.128188
626	0.125112
627	0.125152
628	0.125913
629	0.141193
630	0.127426
631	0.125853
632	0.126665
633	0.127116
634	0.1247
635	0.128649
636	0.131373
637	0.136253
638	0.125583
639	0.132126
640	0.196249
641	0.174768
642	0.186069
643	0.190708
644	0.163526
645	0.195067
646	0.168336
647	0.162224
648	0.161783
649	0.198774
650	0.179267
651	0.197181
652	0.187231
653	0.151843
654	0.141283
655	0.121164
656	0.11949
657	0.118048
658	0.138738
659	0.117908
660	0.173645
661	0.164749
662	0.165449
663	0.154769
664	0.120924
665	0.118058
666	0.117496
667	0.121184
668	0.12467
669	0.137074
670	0.117858
671	0.118799
672	0.137666
673	0.118378
674	0.118509
675	0.141353
676	0.276093
677	0.193543
678	0.226468
679	0.189696
680	0.15548
681	0.151983
682	0.152714
683	0.15525
684	0.151171
685	0.149429
686	0.149108
687	0.139519
688	0.116334
689	0.172884
690	0.119241
691	0.137175
692	0.115523
693	0.19134
694	0.191049
695	0.16107
696	0.142595
697	0.123428
698	0.141243
699	0.158917
700	0.16029
701	0.161722
702	0.16039
703	0.19685
704	0.174287
705	0.240113
706	0.167544
707	0.158676
708	0.157805
709	0.16039
710	0.164978
711	0.190738
712	0.170489
713	0.16074
714	0.162955
715	0.1194
716	0.118629
717	0.138317
718	0.11948
719	0.145591
720	0.151362
721	0.154528
722	0.150521
723	0.149238
724	0.148596
725	0.238891
726	0.204825
727	0.459457
728	0.302644
729	0.223522
730	0.142194
731	0.127436
732	0.149668
733	0.123408
734	0.18647
735	0.153155
736	0.159177
737	0.159298
738	0.164176
739	0.162543
740	0.123368
741	0.11927
742	0.118709
743	0.117737
744	0.117617
745	0.117386
746	0.116364
747	0.141883
748	0.19128
749	0.166752
750	0.12473
751	0.121454
752	0.11907
753	0.115964
754	0.151693
755	0.118468
756	0.116164
757	0.141904
758	0.147525
759	0.157123
760	0.154859
761	0.148928
762	0.143186
763	0.1503
764	0.12938
765	0.159558
766	0.246475
767	0.239713
768	0.167704
769	0.163024
770	0.141342
771	0.117527
772	0.117377
773	0.153576
774	0.117366
775	0.118479
776	0.116615
777	0.148456
778	0.131153
779	0.117306
780	0.115924
781	0.115032
782	0.115924
783	0.147905
784	0.135351
785	0.133007
786	0.143116
787	0.153436
788	0.542487
789	0.225385
790	0.199836
791	0.145641
792	0.149528
793	0.145911
794	0.141663
795	0.223631
796	0.26945
797	0.294599
798	0.265703
799	0.140701
800	0.122877
};
\addlegendentry{HPC}

\end{axis}
\end{tikzpicture}

%% file: figures/runtime_boxplot.tex
\begin{tikzpicture}[font=\footnotesize]
  \begin{axis}[
    width=7cm,
    height=3cm,
    scale only axis,
    scaled ticks=false,
    scaled ticks=false,
    xmin=1.05,
    xmax=1.25,
    boxplot/draw direction=x,
    xlabel={Runtime in \si{\milli\second}},
    ylabel={Run},
    ytick={1,2,3},
    yticklabels={1, 2, 3},
    boxplot/box extend=0.8,
  ]
    % Run 10k
    \addplot+[
      mark=*,
      mark options={color=black, scale=0.7},
      fill=Blue,
      boxplot prepared={
        lower whisker=1.086750,
        lower quartile=1.093250,
        median=1.096375,
        upper quartile=1.105125,
        upper whisker=1.122250,
        average=1.0990123
      },
      draw=black
    ] coordinates {
    (1,1.127625)
    (1,1.141375)
    (1, 1.157250)
    (1, 1.1983750)
    };
    %%% Run 1
    \addplot+[
      mark=*,
      mark options={color=black, scale=0.7},
      fill=Bluelight,
      boxplot prepared={
        lower whisker=1.089125,
        lower quartile=1.094625,
        median=1.0970625,
        upper quartile=1.106375,
        upper whisker=1.113375,
        average=1.1003029
      },
      draw=black
    ] coordinates {
    (2, 1.155375)
    (2, 1.199375)
    };
    %%% Run 2
    \addplot+[
      mark=*,
      mark options={color=black, scale=0.7},
      fill=LightGray,
      boxplot prepared={
        lower whisker=1.092750,
        lower quartile=1.096875,
        median=1.098875,
        upper quartile=1.108125,
        upper whisker=1.114125,
        average=1.10203375
      },
      draw=black
    ] coordinates {
    (3, 1.139125)
    (3, 1.205500)
    };
  \end{axis}
\end{tikzpicture}

%% file: figures/trajectory_colors.tex
\definecolor{coral24612474}{RGB}{246,124,74}
\definecolor{coral24813981}{RGB}{248,139,81}
\definecolor{coral24814282}{RGB}{248,142,82}
\definecolor{coral24914784}{RGB}{249,147,84}
\definecolor{crimson2185642}{RGB}{218,56,42}
\definecolor{darkgray176}{RGB}{176,176,176}
\definecolor{darkkhaki154212104}{RGB}{154,212,104}
\definecolor{darkkhaki157213105}{RGB}{157,213,105}
\definecolor{darkkhaki159214105}{RGB}{159,214,105}
\definecolor{darkkhaki162215105}{RGB}{162,215,105}
\definecolor{darkkhaki164216105}{RGB}{164,216,105}
\definecolor{darkkhaki167217106}{RGB}{167,217,106}
\definecolor{darkkhaki171219109}{RGB}{171,219,109}
\definecolor{darkkhaki173220110}{RGB}{173,220,110}
\definecolor{darkkhaki177221113}{RGB}{177,221,113}
\definecolor{darkkhaki179222114}{RGB}{179,222,114}
\definecolor{darkkhaki181223115}{RGB}{181,223,115}
\definecolor{darkkhaki183224117}{RGB}{183,224,117}
\definecolor{darkkhaki185225118}{RGB}{185,225,118}
\definecolor{darkkhaki187226119}{RGB}{187,226,119}
\definecolor{darkkhaki189226120}{RGB}{189,226,120}
\definecolor{darkkhaki191227122}{RGB}{191,227,122}
\definecolor{darkkhaki193228123}{RGB}{193,228,123}
\definecolor{darkseagreen112193100}{RGB}{112,193,100}
\definecolor{darkseagreen114194100}{RGB}{114,194,100}
\definecolor{darkseagreen119196100}{RGB}{119,196,100}
\definecolor{darkseagreen122197101}{RGB}{122,197,101}
\definecolor{darkseagreen124198101}{RGB}{124,198,101}
\definecolor{darkseagreen129201102}{RGB}{129,201,102}
\definecolor{darkseagreen134203102}{RGB}{134,203,102}
\definecolor{darkseagreen137204102}{RGB}{137,204,102}
\definecolor{darkseagreen139205103}{RGB}{139,205,103}
\definecolor{darkseagreen142206103}{RGB}{142,206,103}
\definecolor{darkseagreen144207103}{RGB}{144,207,103}
\definecolor{firebrick166138}{RGB}{166,1,38}
\definecolor{firebrick1862038}{RGB}{186,20,38}
\definecolor{forestgreen010455}{RGB}{0,104,55}
\definecolor{forestgreen1012264}{RGB}{10,122,64}
\definecolor{forestgreen210756}{RGB}{2,107,56}
\definecolor{forestgreen611560}{RGB}{6,115,60}
\definecolor{forestgreen711761}{RGB}{7,117,61}
\definecolor{forestgreen811962}{RGB}{8,119,62}
\definecolor{forestgreen912063}{RGB}{9,120,63}
\definecolor{gray}{RGB}{128,128,128}
\definecolor{khaki195229124}{RGB}{195,229,124}
\definecolor{khaki197230126}{RGB}{197,230,126}
\definecolor{khaki199231127}{RGB}{199,231,127}
\definecolor{khaki201232128}{RGB}{201,232,128}
\definecolor{khaki203232129}{RGB}{203,232,129}
\definecolor{khaki205233131}{RGB}{205,233,131}
\definecolor{khaki207234132}{RGB}{207,234,132}
\definecolor{khaki209235133}{RGB}{209,235,133}
\definecolor{khaki215238137}{RGB}{215,238,137}
\definecolor{khaki217239139}{RGB}{217,239,139}
\definecolor{khaki222241147}{RGB}{222,241,147}
\definecolor{khaki253214130}{RGB}{253,214,130}
\definecolor{khaki253222137}{RGB}{253,222,137}
\definecolor{khaki254225141}{RGB}{254,225,141}
\definecolor{khaki254230149}{RGB}{254,230,149}
\definecolor{khaki254233155}{RGB}{254,233,155}
\definecolor{lemonchiffon252254187}{RGB}{252,254,187}
\definecolor{lemonchiffon254250183}{RGB}{254,250,183}
\definecolor{lemonchiffon254253187}{RGB}{254,253,187}
\definecolor{lightsalmon253198117}{RGB}{253,198,117}
\definecolor{mediumseagreen10218999}{RGB}{102,189,99}
\definecolor{mediumseagreen10419099}{RGB}{104,190,99}
\definecolor{mediumseagreen10719199}{RGB}{107,191,99}
\definecolor{mediumseagreen10919299}{RGB}{109,192,99}
\definecolor{mediumseagreen5716787}{RGB}{57,167,87}
\definecolor{mediumseagreen6016888}{RGB}{60,168,88}
\definecolor{mediumseagreen6317089}{RGB}{63,170,89}
\definecolor{mediumseagreen7217491}{RGB}{72,174,91}
\definecolor{mediumseagreen7817793}{RGB}{78,177,93}
\definecolor{mediumseagreen8117893}{RGB}{81,178,93}
\definecolor{mediumseagreen8718195}{RGB}{87,181,95}
\definecolor{mediumseagreen9018396}{RGB}{90,183,96}
\definecolor{mediumseagreen9318496}{RGB}{93,184,96}
\definecolor{mediumseagreen9618697}{RGB}{96,186,97}
\definecolor{mediumseagreen9918798}{RGB}{99,187,98}
\definecolor{moccasin254244173}{RGB}{254,244,173}
\definecolor{moccasin254245175}{RGB}{254,245,175}
\definecolor{moccasin254247177}{RGB}{254,247,177}
\definecolor{palegoldenrod228244155}{RGB}{228,244,155}
\definecolor{palegoldenrod230244157}{RGB}{230,244,157}
\definecolor{palegoldenrod233245161}{RGB}{233,245,161}
\definecolor{palegoldenrod236247165}{RGB}{236,247,165}
\definecolor{palegoldenrod237247167}{RGB}{237,247,167}
\definecolor{palegoldenrod239248169}{RGB}{239,248,169}
\definecolor{palegoldenrod240249171}{RGB}{240,249,171}
\definecolor{palegoldenrod254239165}{RGB}{254,239,165}
\definecolor{sandybrown25015488}{RGB}{250,154,88}
\definecolor{sandybrown253182104}{RGB}{253,182,104}
\definecolor{sandybrown253190111}{RGB}{253,190,111}
\definecolor{seagreen1513269}{RGB}{15,132,69}
\definecolor{seagreen1713671}{RGB}{17,136,71}
\definecolor{seagreen1813772}{RGB}{18,137,72}
\definecolor{seagreen1913973}{RGB}{19,139,73}
\definecolor{seagreen2014174}{RGB}{20,141,74}
\definecolor{seagreen2114375}{RGB}{21,143,75}
\definecolor{seagreen2214576}{RGB}{22,145,76}
\definecolor{seagreen2414978}{RGB}{24,149,78}
\definecolor{seagreen2515179}{RGB}{25,151,79}
\definecolor{seagreen3015481}{RGB}{30,154,81}
\definecolor{seagreen3315581}{RGB}{33,155,81}
\definecolor{seagreen3615782}{RGB}{36,157,82}
\definecolor{seagreen4516184}{RGB}{45,161,84}
\definecolor{seagreen4816285}{RGB}{48,162,85}
\definecolor{seagreen5116486}{RGB}{51,164,86}